\documentclass[11pt,a4paper]{article}
\usepackage[hyperref]{acl2021}

\usepackage{times}
\usepackage{latexsym}

\usepackage{amsmath,bm}
\usepackage{mathrsfs}
\usepackage{multirow}
\usepackage{CJKutf8}
\usepackage{verbatim}
\usepackage{amssymb}
\usepackage{algorithm}
\usepackage{algcompatible}
\usepackage{caption}
\usepackage{graphicx}
\usepackage{color}
\usepackage{subfigure}
\usepackage{hyperref} 
\usepackage{enumitem}
\usepackage{makecell}
\usepackage{soul}
\usepackage[bottom]{footmisc}

\setul{1pt}{.4pt}

\definecolor{darkred}{HTML}{CA2826}
\definecolor{myblue}{HTML}{2F5597}
\definecolor{darkgreen}{HTML}{548235}

\newcommand{\textmore}[1]{\textcolor{darkred}{\textit{#1}}}
\newcommand{\textreplace}[1]{\textcolor{darkgreen}{\textbf{#1}}}
\newcommand{\textadd}[1]{\textcolor{myblue}{\ul{#1}}}
\newcommand{\textdel}[1]{\textcolor{myblue}{\ul{#1}}}

\newcommand{\textcorrect}[1]{\textcolor{myblue}{\textit{\textbf{#1}}} }

\newcommand{\textunk}[1]{\textit{#1} }

\usepackage{microtype}

\aclfinalcopy 
\def\aclpaperid{1203} 

\newcommand{\dX}{{\mathcal{X}}}
\newcommand{\dY}{{\mathcal{Y}}}

\newcommand{\E}{{\mathbb{E}}}
\DeclareMathOperator*{\argmax}{arg\,max}

\algnewcommand\algorithmicreturn{\textbf{return}}
\algnewcommand\RETURN{\State \algorithmicreturn}%

\title{NAST: A Non-Autoregressive Generator with Word Alignment \\ for Unsupervised Text Style Transfer}

\author{Fei Huang, Zikai Chen, Chen Henry Wu, Qihan Guo, Xiaoyan Zhu, Minlie Huang\footnotemark[1]\\
  The CoAI group, DCST; Institute for Artificial Intelligence;\\
  State Key Lab of Intelligent Technology and Systems;\\
  Beijing National Research Center for Information Science and Technology;\\ 
  Tsinghua University, Beijing 100084, China. \\
  {\small \tt f-huang18@mails.tsinghua.edu.cn}
  \quad {\small \tt natnstart@gmail.com}
  \quad {\small \tt henrychenwu98@gmail.com}\\
  {\small \tt gqh18@mails.tsinghua.edu.cn}
  \quad {\small \tt zxy-dcs@tsinghua.edu.cn}
  \quad {\small \tt aihuang@tsinghua.edu.cn}}

\date{}

\begin{document}

\maketitle

\renewcommand{\thefootnote}{\fnsymbol{footnote}}
\footnotetext[1]{Corresponding author: Minlie Huang.}
\renewcommand{\thefootnote}{\arabic{footnote}}

\begin{abstract}

Autoregressive models have been widely used in unsupervised text style transfer. Despite their success, these models still suffer from the content preservation problem that they usually ignore part of the source sentence and generate some irrelevant words with strong styles. In this paper, we propose a Non-Autoregressive generator for unsupervised text Style Transfer (NAST), which alleviates the problem from two aspects. First, we observe that most words in the transferred sentence can be aligned with related words in the source sentence, so we explicitly model word alignments to suppress irrelevant words. Second, existing models trained with the cycle loss align sentences in two stylistic text spaces, which lacks fine-grained control at the word level. The proposed non-autoregressive generator focuses on the connections between aligned words, which learns the word-level transfer between styles. For experiments, we integrate the proposed generator into two base models and evaluate them on two style transfer tasks. The results show that NAST can significantly improve the overall performance and provide explainable word alignments. Moreover, the non-autoregressive generator achieves over 10x speedups at inference. Our codes are available at \url{https://github.com/thu-coai/NAST}.

\end{abstract}

\addtolength{\textfloatsep}{-1em}
\addtolength{\skip\footins}{-0.3em}

\section{Introduction}
\label{sec:introduction}

Text style transfer aims at changing the text style while preserving the style-irrelevant contents, which has a wide range of applications, e.g., sentiment transfer \cite{crossalign2017shen}, text formalization \cite{GYAFC2018rao}, and author imitation \cite{shakespear2017harsh}. Due to the lack of parallel training data, most works focus on \textit{unsupervised} text style transfer using non-parallel stylistic data.


The cycle consistency loss \cite{cyclegan2017zhu}, a.k.a. the back-translation loss \cite{umt2018lample, multiattr2019lample}, has been widely adopted by unsupervised text style transfer models \cite{styletrans2019dai, latentseq2020he, styins2020yi}. Specifically, the cycle loss minimizes the reconstruction error for the sentence transferred from style $\dX$ to style $\dY$ and then back to $\dX$,
which aligns the sentences in two stylistic text spaces to achieve the transfer and preserve style-irrelevant contents. 
The cycle-loss-based models are trained in an end-to-end fashion, and thus can be easily applied to different datasets.



\begin{figure}[!t]
  \centering
  \includegraphics[width=0.92\linewidth]{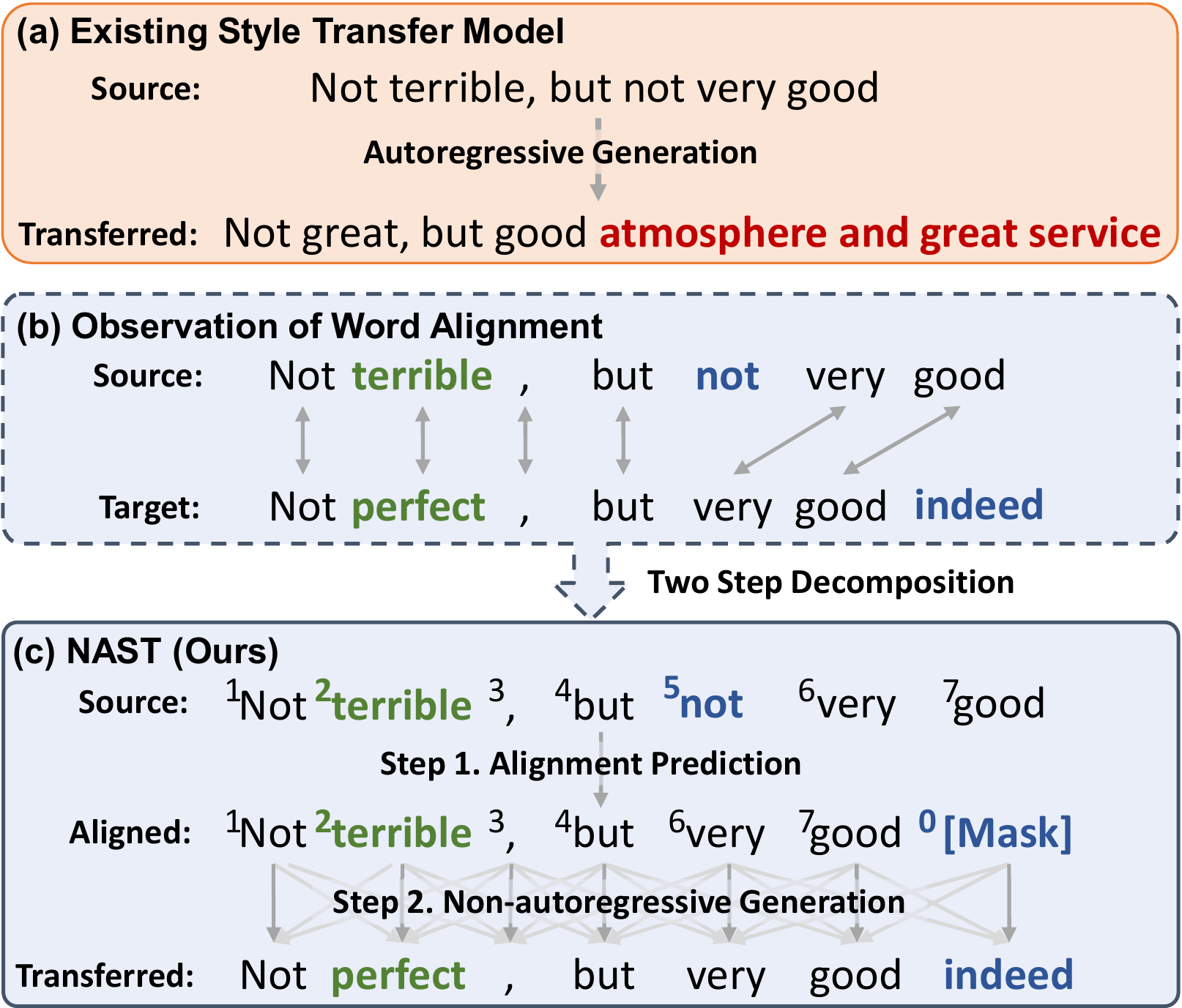}
  \caption{Sentiment transfer examples (negative to positive). (a) Existing models without word alignments may generate \textcolor{darkred}{\textbf{words irrelevant to the source sentence}}. (b) An example of word alignments between the source and target sentences. Arrows connect aligned words (identical or \textcolor{darkgreen}{\textbf{relevant}}), and \textcolor{myblue}{\textbf{blue words are not aligned}}. (c) NAST's generation process. Step 1: generate the index of aligned words. [Mask] is a placeholder for unaligned words. Step 2: generate the transferred sentence non-autoregressively.}
  \label{fig:word-alignment-case}
  \vspace{-0.5em}
\end{figure}

%
%
Although cycle-loss-based models yield promising results,
one of their major failure cases is to replace some part of the source sentence with irrelevant words that have strong styles, as shown in Fig \ref{fig:word-alignment-case}(a).
This problem degrades content preservation and can be alleviated from two perspectives.
%
%
\textbf{First}, we observe that most words in the human-written transferred sentence can be aligned with those in the source sentence. As shown in Fig \ref{fig:word-alignment-case}(b), we can align ``\textit{Not}'' with ``\textit{Not}'', ``\textit{terrible}'' with ``\textit{perfect}'', and leave only a few words unaligned. It shows that humans regard the alignments between words as a key aspect of content preservation, but they are not explicitly modeled by cycle-loss-based models yet. 
%
\textbf{Second}, existing models use the cycle loss to align sentences in two stylistic text spaces, which lacks control at the word level.
For example, in sentiment transfer, ``\textit{tasty}'' should be mapped to ``\textit{awful}'' (because they both depict food tastes) but not ``\textit{expensive}''. 
We utilize a non-autoregressive generator to model the word-level transfer, where the transferred words are predicted based on contextual representations of the aligned source words.
In this paper, we propose a Non-Autoregressive generator for unsupervised Style Transfer (NAST), 
which explicitly models word alignment for better content preservation.
%
%
Specifically, our generation process is decomposed into two steps: first predicting \textit{word alignments} conditioned on the source sentence, and then generating the transferred sentence with a \textit{non-autoregressive} (NAR) decoder.
Modeling word alignments directly suppresses the generation of irrelevant words, and the NAR decoder exploits the word-level transfer.
NAST can be used to replace the autoregressive generators of existing cycle-loss-based models.
In the experiments, we integrate NAST into two base models: StyTrans  \cite{styletrans2019dai} and LatentSeq \cite{latentseq2020he}. Results on two benchmark datasets show that NAST steadily improves the overall performance.
Compared with autoregressive models, NAST greatly accelerates training and inference and provides better optimization of the cycle loss. Moreover, we observe that NAST learns explainable word alignments.
Our contributions are:

\vspace{-0.5em}
\begin{itemize}[leftmargin=1em]
    \setlength{\itemsep}{0ex}
    \setlength{\parskip}{2px}
  
    \item We propose NAST, a Non-Autoregressive generator for unsupervised text Style Transfer. By explicitly modeling word alignments, NAST suppresses irrelevant words and improves content preservation for the cycle-loss-based models. To the best of our knowledge, we are the first to introduce a non-autoregressive generator to an unsupervised generation task.

    \item Experiments show that incorporating NAST in cycle-loss-based models significantly improves the overall performance and the speed of training and inference. In further analysis, we find that NAST provides better optimization of the cycle loss and learns explainable word alignments.
\end{itemize}

\section{Related Work}

\begin{figure*}[!t]
  \centering
  \includegraphics[width=0.97\linewidth]{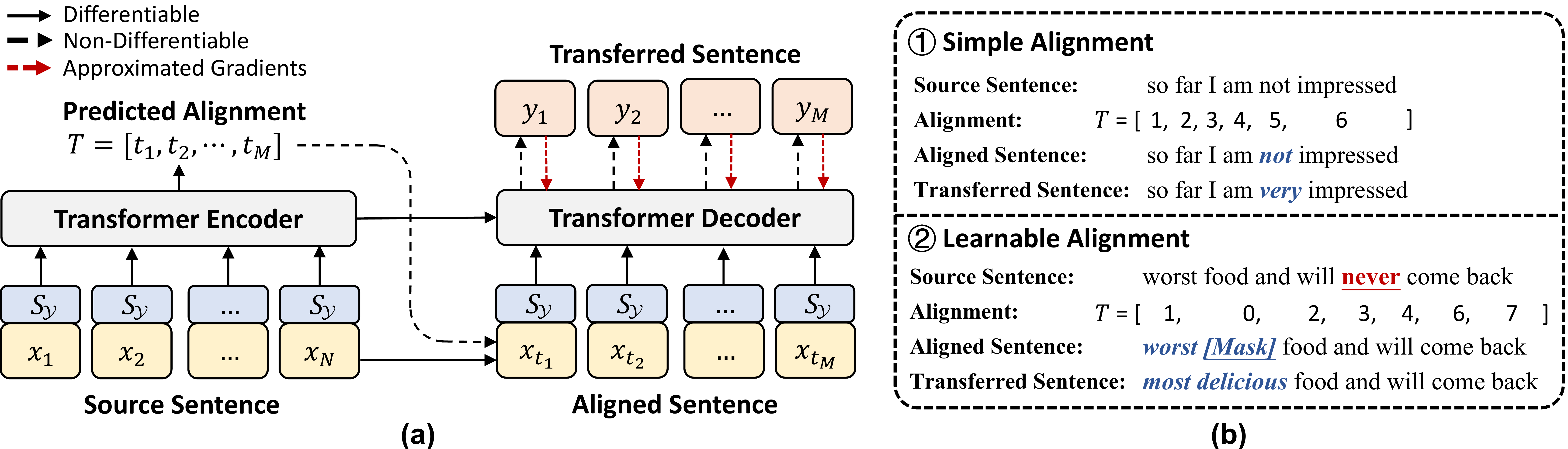}
  \caption{(a) Architecture of NAST transferring $X$ to $Y$. $S_\dY$ is the target style. The NAR decoder generates each word $y_i$ independently. (b) Examples of two alignment prediction strategies. \textbf{Simple Alignment}: each $y_i$ is aligned with $x_i$. \textbf{Learnable Alignment}: a network predicts the alignment, where $t_k=0$ indicates a [Mask] placeholder.}
  \label{fig:model-overview}
  
  
  
  
  \vspace{-1.3em}
\end{figure*}

\noindent \textbf{Unsupervised Text Style Transfer} \ \ 

We categorize style transfer models into three types.
The first type \cite{crossalign2017shen, ARAE2018zhao, lmasdis2018yang, disen2019john} disentangles the style and content representations,  
and then combines the content representations with the target style to generate the transferred sentence.
However, the disentangled representations are limited in capacity and thus hardly scalable for long sentences \cite{styletrans2019dai}.
The second type is the editing-based method \cite{delretgen2018li, hieRL2019wu, maskandinfill2019wu}, which edits the source sentence with several discrete operations. The operations are usually trained separately and then constitute a pipeline. 
%
These methods are highly explainable, but they usually need to locate and replace the stylist words, which hardly applies to complex tasks that require changes in sentence structures.
Although our two-step generation seems similar to a pipeline, NAST is trained in an end-to-end fashion with the cycle loss. All transferred words in NAST are generated, not copied, which is essentially different from these methods.
%
The third type is based on the cycle loss. \citet{stasumt2018zhang, multiattr2019lample} introduce the back translation method into style transfer, where the model is directly trained with the cycle loss after a proper initialization. The following works \cite{styletrans2019dai, dualrl2019luo, latentseq2020he, styins2020yi} further adopt a style loss to improve the style control. 

A recent study \cite{wordsty2020zhou} explores the word-level information for style transfer, which is related to our motivation. However, they focus on word-level style relevance in designing novel objectives, while we focus on modeling word alignments and the non-autoregressive architecture.

\vspace{0.3em}
\noindent \textbf{Non-Autoregressive Generation} \ \ 

Non-AutoRegressive (NAR) generation is first introduced in machine translation for parallel decoding with low latency \cite{nonauto2018gu}.
The NAR generator assumes that each token is generated independently of each other conditioned on the input sentence, which sacrifices the generation quality in exchange for the inference speed.

Most works on NAR generation focus on improving the generation quality while preserving the speed acceleration in machine translation. \citet{nonauto2018gu} find the decoder input is critical to the generation quality. Several works \cite{pnat2019bao, reordernat2019ran} improve the decoder input by aligning source words with target words, which utilize a two-step generation process and inspire the design of NAST.
To our knowledge, only a few works of NAR generation explore applications other than machine translation \cite{natdialog2020han, nat2speech2020peng}.
We are the first to apply NAR generators to an unsupervised text generation task, which surprisingly outperforms autoregressive models in transfer quality besides the acceleration.

\section{Methods}
\label{sec:methods}

\addtolength{\abovedisplayskip}{-1em} 
\addtolength{\belowdisplayskip}{-0.3em} 



In this paper, we formulate the unsupervised text style transfer as follows:
for two non-parallel corpora with styles $\dX$ and $\dY$ respectively, the task aims at training a style transfer model $G$. The model learns the transfer of two directions, $\dX \rightarrow \dY$ and $\dY \rightarrow \dX$, which can be denoted as $P_{G_\dY}(Y|X)$ and $P_{G_\dX}(X|Y)$, respectively.

\subsection{NAST}

NAST is a non-autoregressive generator based on the observation of the word alignment: in style transfer tasks, most generated words can be aligned with the source words, where each pair of the aligned words is either identical or highly relevant. 
%
For simplicity, we only describe $G_\dY$, where $G_\dX$ shares the architecture and parameters except style embeddings.
Given the source sentence $X = [ x_1, x_2, \cdots, x_N ]$, the generation process of NAST is decomposed into two steps: predicting the alignment $T = [t_1, t_2, \cdots, t_M ]$, and then generating the transferred sentence $Y = [ y_1, y_2, \cdots, y_M ]$.
When $1 \leq t_i \leq N$, the generated word $y_{i}$ is aligned with the source word $x_{t_{i}}$. Otherwise, $y_i$ is not aligned with any source word, where we set $t_i$ to $0$ and fill $x_{t_i}$ with a [Mask] placeholder.
%
Formally, we regard $T$ as a latent variable, and the generation probability is formulated as \vspace{0.3em}
\par\nobreak
{
\small
\begin{align}
    P_{G_\dY}(Y|X) = \sum_{T} P_{G_\dY}(Y|X, T) P_{G_\dY}(T|X) \label{eqn:alignment_decomposition},
\end{align}
}\par\noindent
where $P_{G_\dY}(T|X)$ and $P_{G_\dY}(Y|X, T)$ are modeled

\noindent by an alignment predictor and a non-autoregressive decoder, respectively, as shown in Fig \ref{fig:model-overview}.

\subsubsection{Alignment Predictor}

The alignment predictor predicts the target length $M$ and the alignment $T$ conditioned on the source sentence $X$. We utilize a Transformer \cite{transformer2017vaswani} to encode the source sentence and then explore two alternative strategies to predict $T$.

\vspace{0.2em}
\noindent\textbf{Simple Alignment.}\ \ 
Simple Alignment assumes that the source and target sentences have the same length, and each generated word $y_i$ is exactly aligned with the source word $x_i$. Formally,
\par\nobreak
{
\small
\begin{align}
    P_{G_\dY}(T|X) = \mathbb{I}[M=N] \prod_{i=1}^{M} \mathbb{I}[t_i=i], \notag
\end{align}
}\par\noindent
where $\mathbb{I}[\cdot]$ is the indicator function.
A similar strategy has been adopted by editing-based methods \cite{maskandinfill2019wu, lexicalpipeline2020helbig}, where they simply replace several words in the source sentence. Although this strategy cannot alter the sentence length, it empirically works well on simple tasks, such as sentiment transfer.

\vspace{0.2em}
\noindent\textbf{Learnable Alignment.}\ \ 
Inspired by \citet{reordernat2019ran, pnat2019bao}, we utilize a pointer network \cite{pointernet2015vinyals} on top of the encoder, which predicts the alignment $T$: 
\par\nobreak
{
\small
\begin{align}
    P_{G_\dY}(T|X) = \prod_{i=1}^{M} P_{G_\dY}(t_i|X, t_{<i}). \notag
\end{align}
}\par\noindent
%
The pointer network is essentially an autoregressive generator, but it only generates the alignment $t_i$ pointing to a source word.

\subsubsection{Non-autoregressive Decoder}

The non-autoregressive decoder \cite{nonauto2018gu} is a Transformer that generates each word independently. Formally, we have
\par\nobreak
{
\small
\begin{align}
    P_{G_\dY}(Y|X, T) = \prod_{i=1}^{M} P_{G_\dY}(y_i|X, T).
\end{align}
}\par\noindent
The Transformer decoder takes the aligned sentence $[ x_{t_1}, x_{t_2}, \cdots, x_{t_M} ]$ and the target style embedding $S_\dY$ as inputs. It also contains attention connections from the Transformer encoder.

\subsubsection{Training}
\label{sec:training}

NAST is a generator that can be integrated into existing cycle-loss-based models. These models mainly utilize three losses, and the overall objective $\mathcal{L}$ is defined as $\alpha L_{self} + \beta L_{sty} + \gamma L_{cyc}$, where $\alpha, \beta, \gamma$ are hyper-parameters. 
The \textbf{self-reconstruction loss} $L_{self}$ aims at recovering sentences of both styles from their corrupted versions:
\par\nobreak
{
\small
\begin{align}
    \mathcal{L}_{self} &= -\E_{X \sim P_{\dX}} \left[ \log P_{G_\dX}(X | \widetilde{X}) \right] - \notag \\
    & \quad\quad\quad\quad \E_{Y \sim P_{\dY}} \left[ \log P_{G_\dY}(Y | \widetilde{Y}) \right], \label{eqn:slf_loss}
\end{align}
}\par\noindent
where $\widetilde{X}$ and $\widetilde{Y}$ 
are constructed by word dropout, insertion, and masking \cite{multiattr2019lample}, and $P_\dX$ and $P_\dY$ are the data distributions of two styles.
%
The \textbf{style loss} $L_{sty}$ is used to guide the style of generated sentences, which has various designs by existing works, e.g., adopting a style discriminator \cite{styletrans2019dai} or a language model \cite{latentseq2020he}. In our implementation, the style loss is determined by the base model. We simply present a general formulation:
\par\nobreak
{
\small
\begin{align}
    \mathcal{L}_{sty} &= -\E_{X \sim P_{\dX}} \left[ F(G_{\dY}(X), \dY) \right] - \notag \\
    & \quad\quad\quad\quad \E_{Y \sim P_{\dY}} \left[ F(G_{\dX}(Y), \dX) \right],  \label{eqn:sty_loss}
\end{align}
}\par\noindent
where $F(X, \dX)$ indicates a score that shows to which extent the sentence $X$ has the style $\dX$, and $G_{\dY}(X)$ is the generated sentence sampled from $ P_{G_\dY}(Y|X)$ in two steps:
%
%
$T_{\dY}(X) \sim P_{G_\dY}(T|X)$, $G_{\dY}(X) \sim P_{G_\dY}(Y | X, T_{\dY}(X))$.
At last, the \textbf{cycle loss} $L_{cyc}$ is formulated as
\par\nobreak
{
\small
\begin{align}
    \mathcal{L}_{cyc} &= -\E_{X \sim P_{\dX}} \left[ \log P_{G_\dX}(X | G_{\dY}(X)) \right] - \notag \\
    & \quad\quad\quad\quad \E_{Y \sim P_{\dY}} \left[ \log P_{G_\dY}(Y | G_{\dX}(Y)) \right].  \label{eqn:cyc_loss}
\end{align}
}\par\noindent

However, there still exist two obstacles in optimization.
\textbf{Firstly}, because of the non-differentiable problem, we cannot back-propagate the gradients through the discrete text $G_{\dY}(X)$ in Eq.(\ref{eqn:sty_loss})(\ref{eqn:cyc_loss}). As a common workaround, we adopt the Gumbel-Softmax trick \cite{gumbel2017} to approximate the gradients. 
Therefore, the gradients from $G_{\dY}(X)$ can be back-propagated through the decoder output (Fig \ref{fig:model-overview}(a)). However, the alignment $T_{\dY}(X)$ is remained discrete and non-differentiable, where we simply stop the gradients\footnote{As a result, the alignment predictor (for Learnable Alignment) is not optimized following the gradients from $G_{\dY}(X)$, but with a pseudo label introduced later.}.

\textbf{Secondly}, the losses in Eq.(\ref{eqn:slf_loss})(\ref{eqn:cyc_loss}) are intractable for NAST because the generation probability, e.g. $P_{G_\dY}(Y | X)$, is summed over all alignments as defined in Eq.(\ref{eqn:alignment_decomposition}). We provide solutions for the two alignment strategies separately.

\noindent\textbf{For Simple Alignment.}\ \ 
There is only one valid alignment between $X$ and $Y$, so the generation probability is tractable as
\par\nobreak
{
\small
\begin{gather}
    \log P_{G_\dY}(Y | X) = \log P_{G_\dY}(Y | X, T^*), \notag \\
    \text{where \ } T^* = \argmax_T P_{G_\dY}(T|X) = [1, 2, \ldots, N]. \notag
\end{gather}
}\par\noindent

\noindent\textbf{For Learnable Alignment.}\ \ 
Inspired by \citet{pnat2019bao}, we introduce a heuristic rule to obtain a pseudo alignment $T^*$:
\par\nobreak
{
\small
\begin{gather}
    T^* = \argmax_T \sum_{i=1}^{M} cos(e(y_i), e(x_{t_i})) \notag \\
    s.t. \quad t_i = 0\ \ \text{or}\ \ t_i > t_j\quad \text{for}\ \ \forall\ 1 \leq j < i \leq M, \notag
\end{gather}
}\par\noindent
where $e(\cdot)$ indicates the word embeddings.
We can obtain the pseudo alignment by dynamic programming, and the details are presented in Appendix \ref{sec:app-algorithm}.
In the pseudo alignment, most words in $Y$ are aligned with identical or highly relevant words in $X$, which can be used as a good label to supervise our model.
Next, we derive a tractable lower bound for the generation probability:
\par\nobreak
{
\small
\begin{gather}
    \log P_{G_\dY}(Y | X) \geq \log P_{G_\dY}(Y | X, T^*) + \log P_{G_\dY}(T^* | X). \label{eqn:learnable_decomp}
\end{gather}
}\par\noindent
On the right side, the first term trains the NAR decoder, and the second term trains the alignment predictor.
By substituting Eq.(\ref{eqn:learnable_decomp}) into Eq.(\ref{eqn:slf_loss})(\ref{eqn:cyc_loss}), we turn to optimize the upper bounds instead of the original intractable losses. The detailed training algorithm is shown in Appendix \ref{sec:app-algorithm}.

\subsection{Discussions}
\label{sec:discussion}

\begin{figure}[!t]
  \centering
  \includegraphics[width=0.95\linewidth]{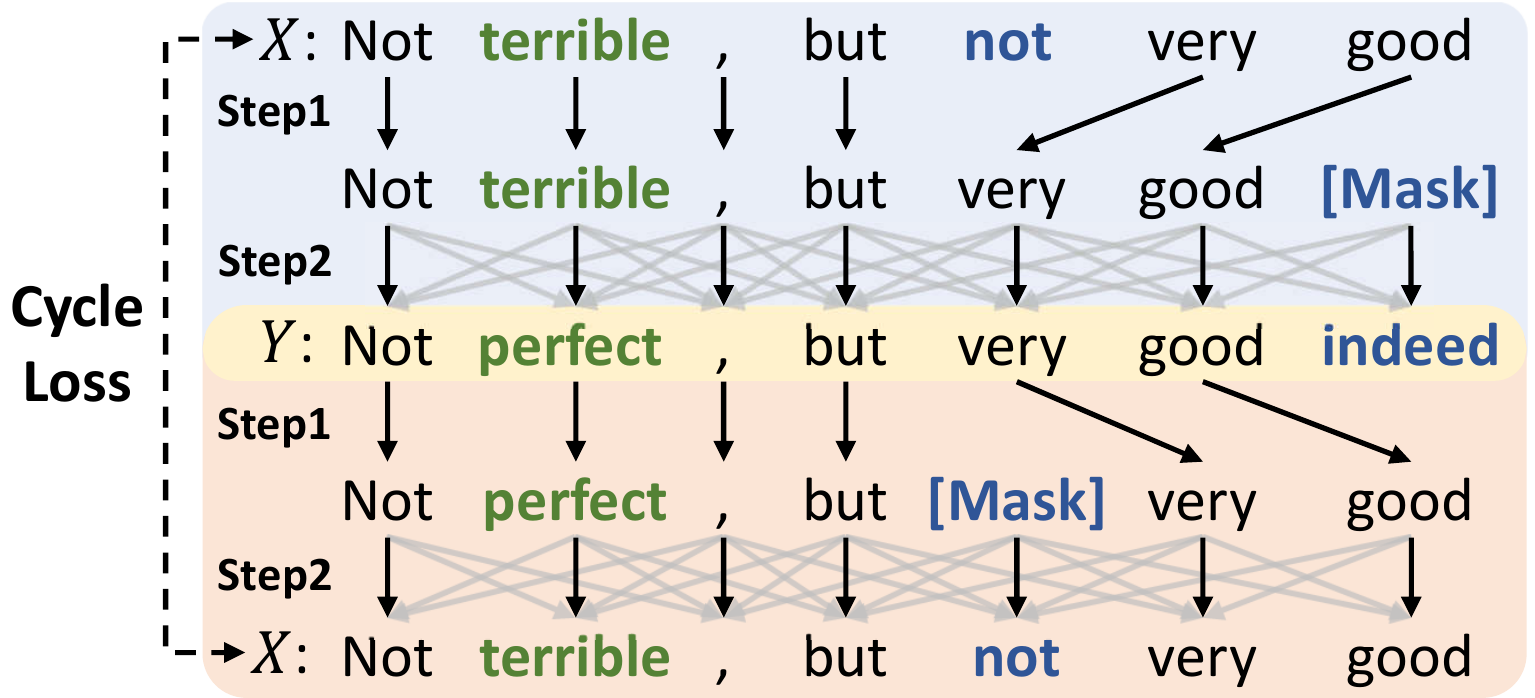}
  \caption{Connections of NAST in the cycle loss with the encoder omitted. The word alignments (step 1) and the residual connections (step 2) are in black.}
  \label{fig:residual}
  \vspace{-0.5em}
\end{figure}

\noindent\textbf{Residual Connections and Multi-head Attention.}
The aligned words in NAST are directly connected with the residual connections, and these connections form several chains in the cycle loss optimization, as shown in Fig \ref{fig:residual}.
Most of these chains represent the word-level transfers and reconstructions, e.g., ``\textit{terrible}'' is transferred to ``\textit{perfect}'' and then reconstructs ``\textit{terrible}''. The reconstruction error is a part of the cycle loss, which is optimized to enhance the alignment in the word space.
%
Besides the residual connections, the multi-head attention mechanism is also important for our model. The attention stops NAST from becoming a degenerate word-to-word dictionary and makes it possible to predict the unaligned words from the context.

\noindent\textbf{Exposure Bias in Autoregressive (AR) Models.}\ \ 
Exposure bias \cite{schedulesampling2015bengio} is a notorious problem in the AR generation. %
To obtain $P_{G_\dX}(X | G_{\dY}(X))$ in the cycle loss, AR generators predict each word of $X$ based on the ground-truth prefix, which is an easy task even without information from $G_{\mathcal{Y}}(X)$.
As a result, in inference, the model may fail in preserving the sentence meaning as it is trained to focus on its generated prefix.
In contrast, NAST focuses on the source sentence since the ground-truth prefix is not given, which suppresses the problem of generating irrelevant words and improves content preservation.
Moreover, the training and test are consistent in NAST\footnote{The claim only applies to NAST with Simple Alignment, because the pseudo alignment used in Learnable Alignment breaks the consistency.}, which alleviates the exposure bias problem.

\section{Experiments}

\subsection{Experiment Settings}

We conduct experiments on two style transfer tasks.

\vspace{0.3em}
\noindent \textbf{Sentiment Transfer.} We use the YELP dataset \cite{delretgen2018li}, which consists of two non-parallel corpora with positive and negative sentiments. For each sentence in the test set, multiple human references are provided by \citet{dualrl2019luo}.

\noindent \textbf{Text Formalization.} We use the family and relationship domain of the GYAFC dataset \cite{GYAFC2018rao}, which consists of paired corpora for formal and informal sentences. We do not use the paired data to supervise training.

\vspace{0.3em}
We utilize several SOTA models as baselines, which include CrossAlign \cite{crossalign2017shen}, DelRetrie \cite{delretgen2018li}, Disent \cite{disen2019john}, StyIns \cite{styins2020yi}, StyTrans \cite{styletrans2019dai}, and LatentSeq \cite{latentseq2020he}.
Our models are modified based on StyTrans and LatentSeq, where we replace their generators with NAST. For StyTrans, NAST adopts a Transformer of the same architecture as the original implementation. However, LatentSeq utilizes an LSTM generator. For a fair comparison, we first incorporate LatentSeq with a vanilla Transformer generator and then replace the generator with NAST of the same architecture.
In inference, we use the greedy decoding strategy, i.e., we choose the top-1 candidate at each step in alignment prediction and sentence generation.
More details are presented in Appendix \ref{sec:app-settings}.

\subsection{Automatic Evaluation}
\label{subsec:automatic-evaluation}

\begin{table*} [!tp]
\centering
\small
\resizebox{0.85\linewidth}{!}{
\setlength{\tabcolsep}{2mm}{
\begin{tabular}{l|rccccc|rccccc}
\hline
 &\multicolumn{6}{c|}{\bf Yelp} &\multicolumn{6}{c}{\bf GYAFC}  \\
\bf Model & \bf PPL & \bf Acc & \bf SelfB & \bf RefB & \bf G2 & \bf H2 & \bf PPL & \bf Acc & \bf SelfB &  \bf RefB & \bf G2 & \bf H2 \\
\hline
CrossAlign \cite{crossalign2017shen} & 105 & 74.0 & 20.3 & 17.9 & 31.8 & 28.8 & 47 & 63.8 & 2.3 & 3.2 & 14.1 & 6.1 \\
DelRetrie \cite{delretgen2018li} & 94 & 88.7 & 36.8 & 31.1 & 52.5 & 46.0 & 101 & 58.2 & 32.3 & 20.8 & 34.2 & 29.8\\
Disent \cite{disen2019john} & \underline{27} & \underline{92.2} & 8.3 & 13.8 & 35.6 & 24.0 & \underline{27} & 68.4 & 4.8 & 8.0 & 23.4 & 14.4 \\
DualRL \cite{dualrl2019luo} & 73 & 88.6 & 59.0 & 55.2 & 68.6 & 67.0 & 91 & 58.9 & 50.1 & 40.3 & 43.9 & 39.2\\
StyIns \cite{styins2020yi} & 98 & 91.5 & 53.2 & 49.0 & 66.9 & 63.7 & 72 & 65.6 & 62.6 & \underline{45.5} & 52.6 & 50.0\\
\hline
StyTrans \cite{styletrans2019dai} & 136 & \bf{90.4} & 53.3 & 48.6 & 66.2 & 63.1 & 124 & 67.1 &  59.7 & 41.9 & 50.4 & 46.8\\
\bf + NAST (Simple) & 117 & 88.9 & \bf{63.3}** & \bf{55.9}** & \bf \underline{70.4}** & \bf \underline{68.5}** & 130 & 67.6 & \bf \underline{63.7}* & 41.6 & 50.8 & 47.4\\
\bf + NAST (Learnable) & \bf{112}* & 87.4 & 62.0** & 54.6** & 69.0** & 67.1** & \bf{119} & \bf \underline{72.9}* & 61.6 & \bf 42.8 & \bf \underline{53.6}** & \bf \underline{49.9}**\\
\hline
LatentSeq \cite{latentseq2020he} & 55 & \bf 84.5 & 49.4 & 47.3 & 62.6 & 60.5 & 47 & 55.3 & \bf 57.8 & 38.5 & 44.1 & 42.5\\
LatentSeq w/ Transformer & \bf{42} & \bf 84.6 & 48.8 & 47.1 & 63.0 & 60.4 & \bf{38} & 58.1 & 54.3 & 35.3 & 45.1 & 43.5 \\
\bf + NAST (Simple) & 73 & 81.2 & 65.2** & 57.6** & \bf{68.1}** & \bf{66.9}** & 56 & 60.4* & 57.0 & 38.2 & 47.4* & 45.6*\\
\bf + NAST (Learnable) & 70 & 79.6 & \bf \underline{65.5}** & \bf \underline{58.0}** & 67.7** & 66.7** & 53 & \bf{64.1}** & 57.0 & \bf 39.2 & \bf{49.0}** & \bf{46.6}*\\
\hline
\end{tabular}
}
}
\caption{Automatic evaluation results. \textbf{Simple} and \textbf{Learnable} indicate two alignment strategies. All values are averaged on two transfer directions. \textbf{Bold} denotes the best results for each base model, and \underline{underline} denotes the best results in all models. * and ** indicate significant improvements over StyTrans or LatentSeq ($p < 0.05$ and $p < 0.01$ in t-test).}
\label{tab:main_result}
\vspace{-1.5em}
\end{table*}

Following \citet{dualrl2019luo, styletrans2019dai}, we utilize a pretrained classifier to evaluate the style accuracy (Acc), and adopt the BLEU-4 score comparing generated sentences with the source sentences (SelfB) or with the references (RefB) to evaluate content preservation. 
The classifier based on RoBERTa-base \cite{RoBERTa2019liu} achieves an accuracy of 97.6\% and 90.1\% on YELP and GYAFC, respectively.
For each transfer direction, we calculate the geometric and harmonic mean of Acc and RefB and then report the average on two directions as G2 and H2, respectively. We further report the perplexity (PPL) of transferred sentences, which is evaluated by GPT2-base \cite{gpt22019} fine-tuned on the training set. 

\phantomsection
\label{txt:task-difference}
The results are shown in Table \ref{tab:main_result}. 
%
Compared with StyTrans and LatentSeq, NAST exhibits stable performance gains of G2 and H2 on both datasets. 
On the Yelp dataset, NAST remarkably improves content preservation (at least 6 points with RefB) but suffers a slight decline in Acc. We find that NAST can suppress irrelevant words with strong styles, which possibly leads to the decline in Acc.
On the GYAFC dataset, NAST outperforms the base models mainly in Acc instead of RefB, which is affected by model selection strategies with the Acc-RefB trade-off. In Table \ref{tab:main_result}, we choose the best model based on G2. A more comprehensive comparisons with trade-off curves will be discussed in the next section.

In terms of the alignment strategies, Learnable Alignment outperforms Simple Alignment on GYAFC, but there is no significant difference on Yelp.
We suppose that the sentiment transfer task is more straightforward than the text formalization, where the model can achieve a good transfer performance on Yelp without changing sentence structures.

Compared with all baselines, our best models set new SOTA results on two datasets in the overall performance of the transfer accuracy and content preservation (i.e., G2 and H2).

\begin{figure}[!t]
  \centering
  \includegraphics[width=\linewidth]{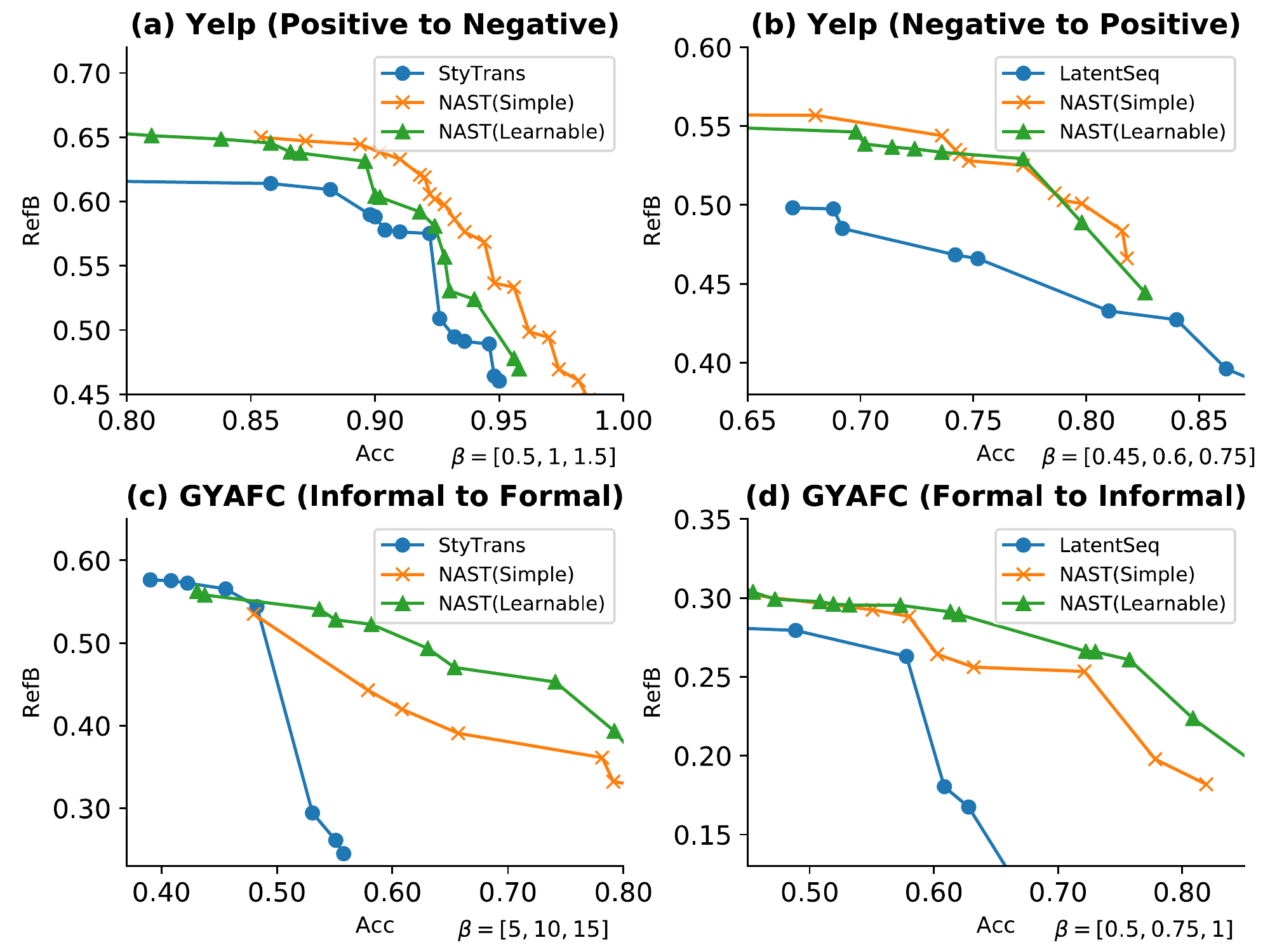}
  \caption{Trade-off curves between style control (Acc) and content preservation (RefB). (a)(c) use StyTrans as the base model, (b)(d) use LatentSeq as the base model. Each curve contains points from three runs with different style loss coefficients $\beta$, whose values for NAST are presented under sub-figures.}
  \label{fig:trade-off}
  \vspace{-0.5em}
\end{figure}

\vspace{0.3em}
\noindent\textbf{Trade-Off Curves.}\ \ 
To investigate the trade-off between style control (shown by Acc) and content preservation (shown by RefB), we follow \citet{eval2018fu} and evaluate the models with different hyper-parameters.
To be specific, we select three different style loss coefficients $\beta$ around the best value.
Please see Appendix \ref{sec:app-hyper} for the search range and other details.
%
Since the trade-off varies through the training, we evaluate the models and collect data points at every epoch. It is different from \citet{eval2018fu}, who only plot the metrics of the best model in each run.  
The curves are shown in Fig \ref{fig:trade-off}, where we only keep the outermost points of each model and remove the points dominated by at least one other point in both Acc and RefB.

The curves of NAST are generally above those of the base models, indicating that NAST achieves better content preservation when the style accuracy is kept at a similar level. In Fig \ref{fig:trade-off} (c)(d), we find that the base model's RefB drops rapidly after Acc exceeds a certain value, which indicates that the cycle loss fails to preserve the sentence-level alignment, thereby leading to model collapse. By contrast, NAST largely alleviates the issue of model collapse. Moreover, we find that Learnable Alignment outperforms Simple Alignment on GYAFC, but performs equally or slightly worse on Yelp, due to the task differences discussed above.

\noindent\textbf{Training \& Inference Speed.}\ \ 
Thanks to the parallel decoding of the NAR generator, NAST accelerates the model training and inference as shown in Table \ref{tab:latency_result}. For a fair comparison, NAST and the corresponding base model utilize the same Transformer architecture. The computation devices are detailed in Appendix \ref{sec:app-devices}.

\begin{table} [!tp]
\centering
\small
\resizebox{1\linewidth}{!}{
\setlength{\tabcolsep}{1mm}{
\begin{tabular}{l|lll}
\hline
 & \bf \#Param &\bf  Train (ms) & \bf Inference (ms) \\
\hline
StyTrans & 31.1M & 857 (1.0x) & 249 (1.0x)\\
\bf + NAST (Simple) & 31.1M & \bf 201 (4.3x) & \bf 8 (31.1x)\\
\bf + NAST (Learnable) & 32.4M & 339 (2.5x) & 71 (3.5x)\\
\hline
LatentSeq w/ Trans. & 21.2M & 1282 (1.0x) & 266 (1.0x)\\
\bf + NAST (Simple) & 21.2M & \bf 714 (1.8x) & \bf 23 (11.6x)\\
\bf + NAST (Learnable) & 22.4M & 761 (1.7x) & 125 (2.1x)\\
\hline
\end{tabular}
}
}
\caption{Parameter size and the training and inference latency on GYAFC. The speedup of training LatentSeq is less significant, because the bottleneck is a language model used in the style loss, costing about 487ms.}
\label{tab:latency_result}
\vspace{-0.6em}
\end{table}

\subsection{Human Evaluation}

We follow \citet{delretgen2018li} and conduct human evaluation experiments on the Yelp dataset. In addition to NAST and the base models, we choose three baselines with the highest G2. For each model, we sample 100 sentences (50 in each transfer direction), and 900 sentences are evaluated in total. For each sentence, three annotators are asked to rate from 1 (worst) to 5 (best) for fluency, style control, and content preservation. 

The human evaluation results are shown in Table \ref{tab:human_result}. Similar to the automatic evaluation results, NAST improves content preservation significantly. Moreover, we find that Learnable Alignment outperforms Simple Alignment in terms of fluency. It can be partially attributed to the fact that Learnable Alignment, which is able to remove or add words, is more flexible in generation.

\subsection{Ablation Study}

\begin{table} [!t]
\centering
\small
\setlength{\tabcolsep}{2mm}{
\begin{tabular}{l|cccc}
\hline
\bf Model & \bf Fluency & \bf Style & \bf Content \\ 
\hline
DelRetrie & 3.87 & 3.90 & 3.05 \\
DualRL & 4.38 & \underline{4.25} & 4.24\\
StyIns & 4.25 & 4.00 & 4.11\\
\hline
StyTrans & 4.24 & \textbf{3.91} & 4.16\\
\bf + NAST(Simple) & 4.34 & 3.87 & \textbf{4.41}**\\
\bf + NAST(Learnable) & \textbf{4.39}* & 3.87 & 4.38**\\
\hline
LatentSeq & 4.53 & 3.92 & 3.59\\
\bf + NAST(Simple) & 4.41 & \textbf{3.93} & 4.43**\\
\bf + NAST(Learnable) & \underline{\textbf{4.57}} & 3.82 & \underline{\textbf{4.48}}**\\
\hline
\end{tabular}
}
\caption{Human evaluation results. \textbf{Bold} denotes the best results for each base model and \ul{underline} denotes the best results among all models. * and ** indicate significant improvements over the base model ($p < 0.05$ and $p < 0.01$ in t-test). The Krippen-dorff’s alpha of human rating is 0.72, indicating acceptable inter-annotator agreement.}
\label{tab:human_result}
\vspace{-0.7em}
\end{table}

\begin{table} [!t]
\centering
\small
\resizebox{1\linewidth}{!}{
\setlength{\tabcolsep}{1mm}{
\begin{tabular}{l|ccc|ccc}
\hline
&\multicolumn{3}{c|}{\bf Yelp} &\multicolumn{3}{c}{\bf GYAFC}  \\
\bf Model & \bf Acc &  \bf RefB & \bf G2 & \bf Acc & \bf RefB & \bf G2\\
\hline
LatentSeq(Trans.) & \bf 84.6  & 47.1 & 63.0  & 58.1 & 35.3 & 45.1\\
NAST(Simple) &  81.3  & 57.4 & \bf 68.1  & 58.3 & 39.3 & \bf47.5\\
\hline
\ \ w/o Aligned Sent. & 73.2  & 44.1 & 56.8  & 54.1 & 34.4 & 42.8\\
\ \ w/o Multi-head Attn. & 57.0  & \bf 62.8 & 59.6 & 22.0 & \bf 41.1 & 29.7\\
\hline
\ \ w/ Soft-Embedding & 44.3 & 44.6 & 43.6 & 63.4 & 26.0 & 40.3\\
\ \ w/ Stop-Gradient & 80.5 & 50.1 & 63.5 & \bf 64.2 & 26.3 & 40.6\\
\hline
\end{tabular}
}
}
\caption{Ablation study of NAR decoder and gradient approximation methods. The base model is LatentSeq.}
\label{tab:nar_ablation_result}
\vspace{-0.6em}
\end{table}

\noindent\textbf{NAR decoder.}\ \ 
Although NAST with Simple Align\-ment has a simple, straightforward design, it works surprisingly well compared with an AR generator. 
%
%
We conduct an ablation study to investigate the impact of different components in the NAR decoder.
First, we remove the \textit{aligned sentence} from the decoder input. Specifically, the decoder input is the positional encodings without the word embeddings. Second, we remove the \textit{multi-head attention} in the decoder, and thus each output word is solely conditioned on its aligned word.

The results are shown in Table \ref{tab:nar_ablation_result}.
After we remove the aligned sentence, the performance drops but still remains comparable. It shows that the multi-head attention over the source sentence learns reasonable transfer, while the performance can be largely improved by providing the decoder with the aligned sentence as input.
%
After we remove the multi-head attention, the overall performance drops remarkably, especially on GYAFC. It shows that NAST utilizes multi-head attention to gather sentence-level information, and it is essentially \textit{not} a word-to-word dictionary. Moreover, the contribution of the multi-head attention is larger on GYAFC than on Yelp. It further justifies that text formalization is less straightforward than sentiment transfer since it requires more sentence-level modifications.  
%
%

\noindent\textbf{Gradient Approximation Methods.}\ \ 
The choice of gradient approximation methods is important for tackling the non-differen\-tiable problem. Besides the Gumbel-Softmax trick used in our full model, we try two alternative methods.
1) The Soft-Embedding approximation \cite{styletrans2019dai} multiplies the softmax distribution by the word embedding matrix to get ``soft'' word embeddings.
2) The Stop-Gradient strategy \cite{latentseq2020he} stops the gradient at the decoder output in the cycle loss. However, the style loss requires the output to be differentiable, so we still apply the Gumbel-Softmax trick for the style loss. 
Results in Table \ref{tab:nar_ablation_result} show that the Gumbel-softmax trick outperforms the other methods, so we utilize the Gumbel-Softmax trick for NAST in other experiments.


\begin{table} [!t]
\centering
\small
\resizebox{1\linewidth}{!}{
\setlength{\tabcolsep}{0.5mm}{






\begin{tabular}{ll}
\hline
\multicolumn{2}{c}{\bf Pseudo Alignment in Self-Reconstruction Loss} \\
\hline
S: & that [Mask] \textmore{talk} , if are not happy \textmore{like} but you . \\
P: & that \textreplace{[Mask]} , if \textadd{[Mask]} are not happy but you \textadd{[Mask]} . \\
T: & that \textreplace{is} , if \textadd{others} are not happy but you \textadd{are} . \\
\hline
\multicolumn{2}{c}{\bf Pseudo Alignment in Cycle Loss} \\
\hline
S: & i leave your email on \textmore{exercise} , and see what happens . \\
P: & \textreplace{i} leave your email \textadd{[Mask]} on \textadd{[Mask]} \textreplace{,} and see what happens \textreplace{.} \\
T: & \textreplace{just} leave your email \textadd{loged} on \textadd{accidentally} \textreplace{...} and see what happens \textreplace{!} \\
\hline
\end{tabular}
}
}
\caption{Pseudo alignments on GYAFC. S = source, P = pseudo alignment, T = target. 
\textmore{Unaligned source words}, \textdel{unaligned target words}, and \textreplace{non-identical aligned words} are marked in different colors.
}
\label{tab:case_pseudo}
\vspace{-0.5em}
\end{table}

\noindent\textbf{Learnable Alignment.}\ \ 
According to Eq.(\ref{eqn:slf_loss})(\ref{eqn:cyc_loss})(\ref{eqn:learnable_decomp}), the alignment predictor in Learnable Alignment is supervised by pseudo alignments when optimizing the upper bounds of the self-reconstruction loss and the cycle loss.
For the former, the alignment predictor learns to align the corrupted $\widetilde{X}$ with $X$. For the latter, the alignment predictor learns to align the transferred sentence $G_{\dY}(X)$ with the original $X$. We show two cases in Table \ref{tab:case_pseudo}, where the pseudo alignments are of acceptable quality. 

To investigate the effects of the pseudo alignments supervision, we remove $\log P_{G_\dY}(T^* | X)$ in Eq.(\ref{eqn:learnable_decomp}) for the two losses separately. Results are shown in Table \ref{tab:predictor_ablation_result}. Without the pseudo alignments supervision in the self-reconstruction loss, the model almost degenerates into Simple Alignment,
because keeping the length unchanged is the easiest way to minimize the cycle loss.
Without the pseudo supervision in the cycle loss, Learnable Alignment is slightly weaker than the full model but still outperforms Simple Alignment. 

\subsection{Case Study of Word Alignment}
\label{subsec:analysis-word-alignment}

\begin{table} [!t]
\centering
\small
\setlength{\tabcolsep}{0.8mm}{
\begin{tabular}{l|ccccc}
\hline
\bf Model & \bf Acc & \bf \bf RefB & \bf G2 & \bf $|\Delta|$ & $std(\Delta)$ \\
\hline
NAST(Simple) & 66.5 &  41.6 & 50.4 & 0.00 & 0.00 \\
NAST(Learnable) & \bf {73.0} & \bf 43.5 & \bf 54.2 & 0.80 & 1.26 \\
\ \ w/o Pseudo(Recon)  & 66.1 & 41.5 & 50.3 & 0.02 & 0.28 \\
\ \ w/o Pseudo(Cyc) & 68.5 & 42.5 & 52.7 & 0.96 & 0.47 \\
\hline
\end{tabular}
}
\caption{Ablation study of NAST with Learnable Alignment on GYAFC.
$\Delta$ is the length difference before and after the transfer. $|\Delta|$ and $std(\Delta)$ indicate the average absolute value and the standard deviation, respectively.
All models use StyTrans as the base model.}
\label{tab:predictor_ablation_result}
\vspace{-0.5em}
\end{table}

\begin{table} [!t]
\centering
\tiny
\resizebox{1\linewidth}{!}{
\setlength{\tabcolsep}{0.5mm}{
\begin{tabular}{ll}
\hline
\multicolumn{2}{c}{\bf Yelp (Positive to Negative)} \\
\hline
Source & love this place and will keep coming back . \\
LatentSeq & \textmore{do n't waste your time} and wo n't be back . \\
StyTrans &  avoid this place and \textmore{will keep coming back} . \\
\bf NAST(Simp.) & \textreplace{skip} this place and will \textreplace{never} coming back . \\
\bf NAST(Lear.) & \textreplace{hate} this place and will \textadd{not} \textreplace{be} coming back . \\
\hline
\multicolumn{2}{c}{\bf Yelp (Negative to Positive)} \\
\hline
Source: &  i did n't even eat it . \\
LatentSeq: &  i always \textmore{love their food and service} . \\
StyTrans: & i love the food eat it . \\
\bf NAST(Simp.): &  i \textreplace{love} \textreplace{it} \textreplace{and} eat it . \\
\bf NAST(Lear.): & i \textadd{definitely} \textreplace{love} \textadd{[DEL]} \textreplace{to} eat it . \\
\hline
\multicolumn{2}{c}{\bf GYAFC (Formal to Informal)} \\
\hline
Source & the world would be happier if men knew what women want . \\
LatentSeq & \textmore{the guy would be mad if they want} what women want . \\
StyTrans & the world would be \textmore{what if thing what girls want girl ur girl want} . \\
\bf NAST(Simp.) & \textreplace{and} world \textreplace{'ll} be happier if men knew what women want . \\
\bf NAST(Lear.) & \textreplace{just} world would be happier \textadd{...} if \textreplace{guys} knew what women want \textdel{[Del]} \\
\hline
\multicolumn{2}{c}{\bf GYAFC (Informal to Formal)} \\
\hline
Source: & i do n't know ! ... i just want the points ... lol \\
LatentSeq: & i do not know . i just want the points . \textmore{however , i am not a good one .} \\
StyTrans: & i do not know ! \\
\bf NAST(Simp.): & i do \textreplace{not} know ! \textreplace{.} i just want the points \textreplace{.} \textreplace{.} \\
\bf NAST(Lear.): & i do \textreplace{not} know ! \textdel{[Del]} i just want the points . \textdel{[Del]} \\
\hline
\end{tabular}
}
}
\caption{Transfer cases. \textmore{Red words} indicate irrelevant phrases or failed transfer in style. \textadd{Non-trivial alignments} and \textreplace{non-identical aligned words} are marked in colors. \textadd{[Del]} indicates the source word is unaligned.}
\label{tab:trans_cases}
\vspace{-0.8em}
\end{table}

We present several transfer cases in Table \ref{tab:trans_cases}. We observe that a major failure mode of the base models is generating irrelevant words. We also observe that NAST achieves better content preservation, and most words in NAST's prediction can be aligned with the source words. 
Focused on the alignment strategies, we observe that the outputs of NAST with Simple Alignment sometimes contain grammar errors (e.g., ``\textit{will never coming back}''), which can be attributed to its limitation of not changing the sentence length. In contrast, we observe that Learnable Alignment can add and remove words at appropriate positions.

To understand the learned \textit{word alignments} and the \textit{word-level transfer}, we count the aligned word pairs based on the prediction of Learnable Alignment. Several cases are presented in Table \ref{tab:align_cases}. 
We observe the aligned word pairs are highly explainable. For example, NAST maps ``\textit{delicious}'' to ``\textit{bland}'' in sentiment transfer and maps ``\textit{guy}'' to ``\textit{man}'' in text formalization.
These cases show that the model can learn fine-grained word-level transfer, where ``\textit{delicious}'' and ``\textit{bland}'' both depict food taste with different styles.
Moreover, NAST with Learnable Alignment learns to add or remove words at reasonable positions, such as adding missing punctuation marks (``\textit{.}'', ``\textit{?}'') and removing redundant words (``\textit{...}'', ``\textit{lol}'') in text formalization. 

\subsection{Analysis of Cycle Loss Optimization}
\label{sec:exp-only-cyc}

\begin{table} [!t]
\centering
\tiny
\resizebox{\linewidth}{!}{
\setlength{\tabcolsep}{0.7mm}{
\begin{tabular}{l|llllllll}
\hline
\multicolumn{5}{c}{\bf NAST(Simple) on Yelp (Negative to Positive)} \\
\hline
\bf Src Word & \multicolumn{4}{c}{\bf Transferred Words} \\
\hline
\textit{helpful} & \textcorrect{weird}(100\%) \\
\textit{fresh} & \textcorrect{tasteless}(61.5\%) & \multicolumn{2}{l}{\textcorrect{overcooked}(38.5\%)} \\
\textit{definitely} & \textcorrect{not}(92.9\%) & \textcorrect{never}(7.1\%) \\
\textit{nice} & \textcorrect{rude}(52.9\%) & \textunk{no}(47.1\%) \\
\textit{best} & \textcorrect{worst}(96.8\%) & \textunk{money}(3.2\%)  \\
\textit{delicious} & \textcorrect{bland}(82.6\%) & \textcorrect{ok}(13.0\%)& \textcorrect{frozen}(4.4\%) \\
\textit{love} & \textcorrect{hate}(63.6\%)& \textunk{ordered}(18.2\%)& \textcorrect{skip}(13.6\%)& \textcorrect{avoid}(4.6\%) \\
\hline
\hline
\multicolumn{5}{c}{\bf NAST(Learnable) on GYAFC (Informal to Formal)} \\
\hline
\bf Src Word &  \multicolumn{4}{c}{\bf Transferred Words} \\
\hline
\textit{'m} & \textcorrect{am}(100\%)\\
\textit{n't} & \textcorrect{not}(98.5\%) & \textunk{n't}(1.5\%)\\
\textit{guy} & \textcorrect{man}(98.4\%)& \textunk{guy}(1.6\%) \\
\textit{u} & \textcorrect{you}(89.2\%)& \textunk{[Del]}(10.8\%) \\
\textit{lol} & \textcorrect{.}(41.7\%)& \textcorrect{[Del]}(41.7\%) & \multicolumn{2}{l}{\textunk{although}(16.7\%)} \\
\textit{...} & \textcorrect{[Del]}(31.3\%)& \textcorrect{,}(27.4\%)& \textcorrect{.}(26.3\%)& and other 7 words \\
\textit{mean} & \textcorrect{believe}(50.0\%) & \textunk{mean}(20.8\%) & \textunk{am}(20.8\%) & and other 2 words \\
\textit{[Mask]} & \textcorrect{.}(55.2\%)& \textunk{a}(12.0\%)& \textcorrect{?}(4.8\%)& and other 28 words \\
\hline
\end{tabular}
}
}
\caption{Cases of aligned word pairs generated by NAST. \textit{[Del]} and \textit{[Mask]} indicate an unaligned source word or an unaligned transferred word, respectively. \textcorrect{Reasonable transfers are in blue.}}
\label{tab:align_cases}
\vspace{-0.5em}
\end{table}

The cycle loss plays a key role in unsupervised style transfer, which achieves style control and content preservation by aligning the sentences in two text spaces.
However, the optimization is not straightforward due to the non-differentiable problem. 
In this section, we study how the cycle loss optimization is affected by the generator architecture and compare a NAR generator with an AR generator\footnote{For a fair comparison, the target sentence length is provided to both models, where the AR generator does not need to predict the EOS token.}.
%
To remove the interference of other losses, we train the model solely with the cycle loss and report the BLEU-4 score of the cycle reconstruction. 
%

The results are shown in Table \ref{tab:cyc_loss_result}. The NAR generator remarkably outperforms the AR generator with all gradient approximation methods. We provide two possible explanations for this observation. 
One reason is that word alignments can help the cycle loss align the text spaces. As discussed in Sec \ref{sec:discussion}, the residual connections directly connect aligned words, which exploits the word-level transfer and reconstruction. Compared with the AR generator that aligns the text spaces at the sentence level, aligning word pairs can be much easier.
%
Another possible reason is the error accumulation caused by the gradient approximation methods.
In each step of the AR generation, the gradient approximation methods are applied to the generated word, and the word is then fed into the model as the next input. As a result, gradients will be approximated multiple times in the back-propagation, and the error brought by the approximation may be accumulated and possibly lead to unstable optimization.

Our analysis provides a perspective to understand how NAST works, and reveals that the generator architecture can deeply affect the optimization in the non-differentiable problem. However, we should be cautious when generalizing the results to other settings. 
We notice inconsistent performance report for the gradient approximation methods \cite{styletrans2019dai, engine2020tu, latentseq2020he}, where the phenomenon needs further study.


\begin{table} [!tp]
\centering
\small
\setlength{\tabcolsep}{1mm}{
\begin{tabular}{l|ccccc}
\hline
& Gumbel-Softmax & Stop-Gradient & Soft-Embedding\\
\hline
NAR & \bf 94.4$\pm$0.4 & \bf 67.6$\pm$4.2 & \bf 33.6$\pm$13.8\\
AR & 84.9$\pm$1.1 & 23.5$\pm$1.1 & 29.9$\pm$15.1\\
\hline
\end{tabular}
}
\caption{BLEU-4 of the cycle reconstruction on the Yelp dataset. The values are reported with mean and standard deviation of three runs with different seeds.}
\label{tab:cyc_loss_result}
\vspace{-0.5em}
\end{table}

\section{Conclusion}

In this paper, we propose NAST, a Non-Autoregre\-ssive generator for unsupervised text Style Transfer. It explicitly models word alignments to suppress irrelevant words and exploits the word-level transfer between different styles. Experiments show that NAST improves the overall performance, provides explainable word alignments, and largely speed up training and inference.

However, we should also notice a potential limitation: NAST relies on the assumption that word alignments exist between the source and target sentences. In a more complicated task that lacks word alignments, NAST may lose its advantage of exploiting the word-level transfer.
In future work, we will improve NAST to tackle noisy word alignments in more challenging datasets and build explainable and faster models for a broader range of unsupervised text generation tasks.

\section*{Acknowledgments}
This work was partly supported by the NSFC projects (Key project with No. 61936010 and regular project with No. 61876096). 
This work was also supported by the Guoqiang Institute of Tsinghua University, with Grant No. 2019GQG1 and 2020GQG0005.

\bibliographystyle{acl_natbib}
\bibliography{acl2021}

\vspace{1em}

\appendix


\begin{algorithm}[!t] 
\caption{DP Algorithm for Pseudo Alignment $DP(X, Y)$\label{alg:dp}}
\footnotesize
\begin{algorithmic}[1] 
\REQUIRE ~~ 
Source sentence $X = [x_1, x_2, \cdots, x_N]$ , \par
\hskip\algorithmicindent\ Target sentence $Y = [y_1, y_2, \cdots, y_M]$.

\State Initialize $f(0, j)=0$ for $\forall\ j=0, 1, \cdots, N$.
\State Initialize $T(0, j)$ as empty lists for $\forall\ j=0, 1, \cdots, N$.
\State Calculate the similarity matrix: \par
\hskip\algorithmicindent $sim_{i, j} = cos(e(y_i), e(x_j))$.
\FOR{$i = 1, 2, \cdots, M$}
\FOR{$j = 0, 1, 2, \cdots, N$}

\State Calculate three choices of $f(i, j)$: \par
\quad $c_1 := f(i-1, j)$ \par
\quad $c_2 := f(i-1, j-1) + sim_{i, j}$\ \ only valid if $j > 0$ \par
\quad $c_3 := f(i, j-1)$\ \ only valid if $j > 0$

\IF{$c_1$ is the maximum choice}
    \STATEx \quad\quad\quad $\triangleright$ \textit{$y_i$ is not aligned.}
    \State $f(i, j) := c_1,\ \ T(i, j) := T(i-1, j) \oplus [0]$
    \STATEx \quad\quad\quad $\triangleright$ \textit{$\oplus$ means list concatenation.}
\ELSIF{$c_2$ is the maximum choice}
    \STATEx \quad\quad\quad $\triangleright$ \textit{$y_i$ is aligned with $x_j$.}
    \State $f(i, j) := c_2,\ \  T(i, j) := T(i-1, j-1) \oplus [j]$
\ELSIF{$c_3$ is the maximum choice}
    \STATEx \quad\quad\quad $\triangleright$ \textit{$y_i$ is aligned with $x_k$, where $k < j$.}
    \State $f(i, j) := c_3,\ \ T(i, j) := T(i, j-1)$
\ENDIF
\ENDFOR
\ENDFOR
\RETURN{} $DP(X, Y) := T(M, N)$
\end{algorithmic}
\end{algorithm}

\begin{algorithm*}[t] 
\caption{Training Algorithm for NAST with Learnable Alignment\label{alg:optimize}}
\footnotesize
\begin{algorithmic}[1] 
\REQUIRE ~~ 
Non-parallel text distribution $P_{\dX}$ and $P_{\dY}$. \quad\quad\quad Max number of batches: $max\_batch$.
\FOR{$iter = 1, 2, \cdots, max\_batch$}
\State Sample $X$ from $P_{\dX}$, $Y$ from $P_{\dY}$.
\State Construct $\widetilde{X}$ and $\widetilde{Y}$ from $X$ and $Y$.
\STATE Use Algorithm \ref{alg:dp} to obtain pseudo alignments for the self-reconstruction loss: \par
\vspace{-0.7em}
{
\small
\begin{equation}
T^*_{X, self} = DP(\widetilde{X}, X), \quad T^*_{Y, self} = DP(\widetilde{Y}, Y). \notag
\end{equation}
}
\vspace{-1.2em}\par
\STATE Calculate the upper bound of the self-reconstruction loss. \par
\vspace{-2em}
{
\small
\begin{align}
    \hat{\mathcal{L}}_{self} &= - \log P_{G_{\dX}}(X|\widetilde{X}, T^*_{X, self}) - \log P_{G_{\dX}}(T^*_{X, self}|\widetilde{X}) \notag \\
    & \quad \quad - \log P_{G_{\dY}}(Y|\widetilde{Y}, T^*_{Y, self}) - \log P_{G_{\dY}}(T^*_{Y, self}|\widetilde{Y}) \notag
\end{align}
}
\vspace{-2em}\par
\State Generate transferred samples with gradient approximation methods: \par
\vspace{-2em}
{
\small
\begin{gather}
T_{\dY} \sim P_{G_{\dY}}(T|X), \quad G_{\dY}(X) \sim P_{G_{\dY}}(Y|X, T_{\dY}). \notag \\
T_{\dX} \sim P_{G_{\dX}}(T|Y), \quad G_{\dX}(Y) \sim P_{G_{\dX}}(X|Y, T_{\dX}). \notag
\end{gather}
}
\vspace{-2em}\par
\State Calculate the style loss. \par
\vspace{-2em}
{
\small
\begin{align}
    \mathcal{L}_{sty} = -\E_{X \sim P_{\dX}} \left[ F(G_{\dY}(X), \dY) \right] - \E_{Y \sim P_{\dY}} \left[ F(G_{\dX}(Y), \dX) \right]. \notag
\end{align}
}
\vspace{-2em}\par
\State Use Algorithm \ref{alg:dp} to obtain pseudo alignments for the cycle loss: \par
\vspace{-2em}
{
\small
\begin{align}
    T^*_{X, cyc} = DP(G_{\dY}(X), X), \quad T^*_{Y, cyc} = DP(G_{\dX}(Y), Y). \notag
\end{align}
}
\vspace{-2em}\par
\State Calculate the upper bound of the cycle loss. \par
\vspace{-2em}
{
\small
\begin{align}
    \hat{\mathcal{L}}_{cyc} &= - \log P_{G_{\dX}}(X|G_{\dY}(X), T^*_{X, cyc}) - \log P_{G_{\dX}}(T^*_{X, cyc}|G_{\dY}(X)) \notag \\
    & \quad \quad - \log P_{G_{\dY}}(Y|G_{\dX}(Y), T^*_{Y, cyc}) - \log P_{G_{\dY}}(T^*_{Y, cyc}|G_{\dX}(Y)) \notag
\end{align}
}
\vspace{-2em}\par
\State Update the model with the loss $\mathcal{L} = \alpha \hat{\mathcal{L}}_{self} + \beta \mathcal{L}_{sty} + \gamma \hat{\mathcal{L}}_{cyc}$.
\ENDFOR
\end{algorithmic}
\end{algorithm*}

\section{Optimization of NAST with Learnable Alignment}
\label{sec:app-algorithm}

Since the generation probability $P_{G_{\dY}}(Y|X)$ is intractable for NAST with Learnable Alignment, we introduce a pseudo alignment $T^*$. For $X=[ x_1, x_2, \cdots, x_N ]$ and $Y=[ y_1, y_2, \cdots, y_M]$, the pseudo alignment $T^*$ is defined by a heuristic rule:
\par\nobreak
{
\small
\begin{align}
    T^* &= \argmax_T V(X, Y) \notag \\
    &= \sum_{i=1}^{M} cos(e(y_i), e(x_{t_i})) \notag \\
    s.t. \quad t_i = 0\ \ &\text{or}\ \ t_j < t_i \leq N \quad \text{for}\ \ \forall\ 1 \leq j < i \leq M, \notag
\end{align}
}\par\noindent
where $e(\cdot)$ indicates the word embeddings. For the unaligned target words, we set $x_{0}$ to a [Mask] placeholder, and set the cosine similarity between the [Mask] placeholder and any other tokens to $0$.

The pseudo alignments are obtained by dynamic programming. We introduce a 2-dim array $f(i, j)$ indicating the maximum value of the objective function $V(X, Y)$ if $Y = [y_1, y_2, \cdots, y_i]$ and $X = [x_1, x_2, \cdots, x_j]$. We further introduce a list $T(i, j)$ that records the best alignment for $f(i, j)$.
The algorithm is presented in Algorithm \ref{alg:dp}. The time complexity is $O(NMd)$, where the bottleneck is calculating the similarity matrix, and $d$ is the dimension of word embeddings.

Based on the pseudo alignments, we derive tractable upper bounds of the losses, which are then optimized to train the model. The full training algorithm is presented in Algorithm \ref{alg:optimize}.

\section{Experiment Settings}

\label{sec:app-settings}

\subsection{Dataset and Evaluation Metrics}

We use the processed datasets provided by \citet{dualrl2019luo}, which can be downloaded at  \href{https://github.com/luofuli/DualRL}{\nolinkurl{https://github.com/luofuli/DualRL}}.
The data statistics are shown in Table \ref{tab:datasets}.

\begin{table} [!tp]
\centering
\small
\resizebox{1\linewidth}{!}{
\setlength{\tabcolsep}{1mm}{
\begin{tabular}{c|cccccc}
\hline
\bf Dataset & \bf Styles & \bf \#Train & \bf \#Valid & \bf \#Test & \bf $|V|$ & \bf Avg Len \\
\hline
\multirow{2}{*}{Yelp} & Neg. & 177k & 2,000 & 500 &  \multirow{2}{*}{9,943} & 9.55 \\
& Pos. & 266k & 2,000 & 500 &  & 8.43 \\
\hline
\multirow{2}{*}{GYAFC} & Inf. & 52k & 2,788 & 1,332 & \multirow{2}{*}{26,790} & 13.06 \\
& For. & 52k & 2,247 & 1,019 & & 12.47\\
\hline
\end{tabular}
}
}
\caption{Data statistics. Average length is calculated on the training set.}
\label{tab:datasets}
\vspace{-0.5em}
\end{table}

The pretrained classifier is implemented based on the \textit{transformers} package\footnote{\href{https://github.com/huggingface/transformers}{\nolinkurl{https://github.com/huggingface/transformers}}}, and the BLEU-4 score is the corpus BLEU implemented in the \textit{nltk} package\footnote{\href{https://www.nltk.org/}{\nolinkurl{https://www.nltk.org/}}}. All results in our paper are evaluated by our implemented codes. The reported results of NAST, StyTrans, and LatentSeq in Figure \ref{tab:main_result} are averaged over three runs with different random seeds.

\subsection{Network Architecture and Hyper-Parameters}
\label{sec:app-hyper}

NAST are implemented based on the base model, StyTrans \cite{styletrans2019dai} and LatentSeq \cite{latentseq2020he}. Their codes can be accessed at \href{https://github.com/fastnlp/style-transformer}{\nolinkurl{https://github.com/fastnlp/style-transformer}} and \href{https://github.com/cindyxinyiwang/deep-latent-sequence-model}{\nolinkurl{https://github.com/cindyxinyiwang/deep-latent-sequence-model}}.

For StyTrans, we follow their implementation and hyper-parameters for the Transformer architecture.
We use $4$ Transformer layers, $4$ attention heads, and $256$-dim hidden cells for both the encoder and the decoder.
For the alignment predictor in Learnable Alignment, we utilize a one-layer Transformer decoder with the same number of attention head and dimension of hidden cells.
Moreover, StyTrans utilizes a discriminator for the style loss, which is built on a 4-layer Transformer encoder with the same architecture above. The discriminator and the generator are trained adversarially. Following their implementation, in each iteration, the discriminator is trained for 10 steps and then the generator is trained for 5 steps.
We utilize the Adam optimizer \cite{adam2015kingma} with the learning rate of $1e-4$ and the batch size of $64$. We choose the gradient approximation method from the Gumbel-Softmax trick \cite{gumbel2017}, the Soft-Embedding approximation \cite{styletrans2019dai}, and the Stop-Gradient strategy \cite{latentseq2020he}.
We select the self-reconstruction loss weight $\alpha$ from $\{0.25, 0.5, 1\}$ and the cycle loss weight $\gamma$ from $\{0.25, 0.5, 1\}$.
We find sometimes the transfer accuracy of one direction can be much higher than that of the other direction, so we separately tune the style loss weights for two directions. To be specific, the overall objective is defined as $\alpha L_{self} + \beta_1 L_{X, sty} + \beta_2 L_{Y, sty} + \gamma L_{cyc}$, where $L_{X, sty} = -\E_{X \sim P_{\dX}} \left[ F(G_{\dY}(X), \dY) \right]$, and $L_{Y, sty} = - \E_{Y \sim P_{\dY}} \left[ F(G_{\dX}(Y), \dX) \right]$. We select $\beta_1$, $\beta_2$ from $\{0.5, 1, 1.5, 3, 5, 10, 15\}$.

For LatentSeq, LSTM is adopted as the generator in their original models. We first replace the LSTM with an autoregressive Transformer as a baseline, which also has $4$ Transformer layers, $4$ attention heads, and $256$-dim hidden cells. Then we replace the autoregressive Transformer with an non-autoregressive Transformer with the same architecture. The alignment predictor is a one-layer Transformer decoder with the same architecture above. However, LatentSeq utilizes a language model for the style loss, which is a 512-dim LSTM. We preserve the implementation of the language model. For optimization, we utilize the RAdam optimizer \cite{radam2020liu} with the learning rate of $1e-3$ and the batch size of $64$. We also try the three gradient approximation methods. We set the cycle-reconstruction loss weight $\gamma = 1$. Following their original implementation, the self-reconstruction weight $\alpha$ is annealed from $1$ to $0$ in the first 60k steps. Similar to NAST on StyleTrans, we tune the the style loss weight on two directions separately, where we select $\beta_1, \beta_2$ from $\{0.15, 0.3, 0.45, 0.6, 0.75\}$ for Yelp and $\{0.5, 0.75, 1, 1.25\}$ for GYAFC.

We manually tune the hyper-parameters and select the best model according the performance on the validation set. For the Yelp dataset, the validation set does not have reference answers, so we use the geometric mean of Acc and SelfB as the overall performance. For the GYAFC dataset, we use the geometric mean of Acc and RefB as the overall performance. 


\subsection{Computing Devices and Running Time}

\label{sec:app-devices}

In our experiment, each run uses approximately 4 Intel Xeon Gold 6226R CPUs at 2.90GHz, and 1 Nvidia Quadro RTX 6000 GPU.
We present the max training step and the training time in Table \ref{tab:training_time}. The best results usually appear in the first half of the training.

\begin{table} [!tp]
\centering
\small
\resizebox{1\linewidth}{!}{
\setlength{\tabcolsep}{1mm}{
\begin{tabular}{l|cccccc}
\hline
\bf Model &\bf Yelp & \bf GYAFC \\
\hline
StyTrans \\
\ \ + NAST(Simple) & 135k steps ($\sim$12h) & 135k steps ($\sim$16h) \\
\ \ + NAST(Learnable) & 135k steps ($\sim$16h) & 135k steps ($\sim$25h) \\
\hline
LatentSeq \\
\ \ + NAST(Simple) & 150k steps ($\sim$18h) & 75k steps ($\sim$16h) \\
\ \ + NAST(Learnable) & 150k steps ($\sim$24h) & 75k steps ($\sim$18h) \\
\hline
\end{tabular}
}
}
\caption{The max training step and the training time of our models.}
\label{tab:training_time}
\vspace{-0.5em}
\end{table}


\section{Transfer Difficulties}

\begin{table} [!t]
\centering
\small
\resizebox{\linewidth}{!}{
\setlength{\tabcolsep}{1mm}{
\begin{tabular}{l|ccc|ccc}
\hline
 \multicolumn{7}{c}{\bf Yelp} \\
 \hline
 &\multicolumn{3}{c|}{\bf Negative to Positive} & \multicolumn{3}{c}{\bf Positive to Negative} \\
\bf Model & \bf Acc & \bf RefB & \bf G2 & \bf Acc &  \bf RefB & \bf G2 \\
\hline
DualRL & 85.4 & 49.6 & 65.1 & 85.8 & 60.8 & 72.3 \\
\hline
StyTrans & \bf 87.5 & 45.4 & 63.0 & \bf 93.1 & 51.6 & 69.3 \\
\bf + NAST (Simple) & 86.2 & \bf 50.1 & \bf 65.7 & 91.6 & \bf 61.6 & \bf 75.1 \\
\bf + NAST (Learnable) & 84.2 & 49.2 & 64.3 & 90.7 & 60.0 & 73.8 \\
\hline
LatentSeq & \bf 82.3 & 42.8 & 59.3 & \bf 86.7 & 51.8 & 67.0 \\
\bf + NAST (Simple) & \bf 82.3  & 49.0 & \bf 63.5 & 80.1 & \bf 66.1 & \bf 72.7 \\
\bf + NAST (Learnable) & 80.5  & \bf 50.2 & \bf 63.5 & 78.7 & 65.8 & 72.0 \\
\hline
\multicolumn{7}{c}{\bf GYAFC} \\
\hline
&\multicolumn{3}{c|}{\bf Formal to Informal} & \multicolumn{3}{c}{\bf Informal to Formal} \\
\bf Model & \bf Acc & \bf RefB & \bf G2 & \bf Acc &  \bf RefB & \bf G2 \\
\hline
DualRL & 86.6 & 24.7 & 46.2 & 31.2 & 55.9 & 41.7 \\
\hline
StyTrans & 87.2 & 28.4 & 49.7 & 46.9 & 55.5 & 51.0 \\
\bf + NAST (Simple) & 85.4 & 28.5 & 49.3 & 49.8 & 54.7 & 52.2 \\
\bf + NAST (Learnable) & \bf 92.1 & \bf 29.9 & \bf 52.5 & \bf 53.7 & \bf 55.7 & \bf 54.7 \\
\hline
LatentSeq & 56.8  & 24.7 & 37.2 & 49.8 & 52.4 & 51.1 \\
\bf + NAST (Simple) & 62.5 & \bf 27.4 & 41.4 & 58.3 & 49.0 & 53.4 \\
\bf + NAST (Learnable) & \bf 69.1 & 25.8 & \bf 42.1 & \bf 59.2 & \bf 52.6 & \bf 55.8 \\
\hline
\end{tabular}
}
}
\caption{Automatic evaluation results on two transfer directions.}
\label{tab:transfer_difficulty}
\vspace{-0.5em}
\end{table}

In Table \ref{tab:transfer_difficulty}, we present the results on two transfer directions of Yelp and GYAFC.
On the Yelp dataset, transferring a negative sentence to a positive one is more difficult than the other direction. One possible reason is that the negative sentences are euphemistic and need changes in sentence structures when transferring to the positive sentiment.
In terms of G2, the text formalization is significantly more difficult than the sentiment transfer. The difficulties of two transfer directions vary across models on GYAFC. Transferring formal sentences to informal ones is harder for DualRL, while the other direction is harder for LatentSeq. 


\section{How to Count Aligned Word Pairs}

In Section \ref{subsec:analysis-word-alignment}, we present cases of the word-level transfer. The aligned word pairs are counted based on the predict alignments $T$, following the rules below:

\vspace{-0.5em}
\begin{itemize}[leftmargin=1em]
    \setlength{\itemsep}{0ex}
    \setlength{\parskip}{2px}  
    \item If $1 \leq t_i \leq N$, we record a pair $x_{t_i} \rightarrow y_i$.
    \item If $t_i = 0$, the transferred word is unaligned, and we record a pair $[Mask] \rightarrow y_i$.
    \item If a source word $x_i$ is not aligned with any transferred word, we record a pair $x_i \rightarrow [Del]$.
\end{itemize}

We then collect all word pairs that have the same source word and calculate the proportion of different transferred words. The results shown in Table \ref{tab:align_cases} is obtained on the test set of two datasets.

\end{document}